\documentclass{article}

\usepackage{microtype}
\usepackage{graphicx}
\usepackage{subfigure}
\usepackage{booktabs} 
\usepackage{amsfonts}       

\usepackage[T1]{fontenc}    
\usepackage{url}            
\usepackage{nicefrac}       
\usepackage{amsmath,amssymb} 
\usepackage[utf8]{inputenc}
\usepackage[english]{babel}

\usepackage{bbm} 
\usepackage{threeparttable} 
\usepackage[round]{natbib} 
\usepackage{hhline}

\DeclareMathOperator*{\argmax}{arg\,max} 
\DeclareMathOperator*{\gelu}{gelu}
\DeclareMathOperator*{\softplus}{softplus}

\usepackage{hyperref}

\usepackage[accepted]{icml2020}

\icmltitlerunning{Self-Attentive Hawkes Process}

\begin{document}

\twocolumn[
\icmltitle{Self-Attentive Hawkes Process}




\begin{icmlauthorlist}
\icmlauthor{Qiang Zhang}{ucl}
\icmlauthor{Aldo Lipani }{ucl}
\icmlauthor{Omer Kirnap}{ucl}
\icmlauthor{Emine Yilmaz }{ucl}
\end{icmlauthorlist}

\icmlaffiliation{ucl}{Center of Artificial Intelligence, University College London, United Kingdom}

\icmlcorrespondingauthor{Qiang Zhang}{qiang.zhang.16@ucl.ac.uk}

\icmlkeywords{Hawkes process, self-attention}

\vskip 0.3in
]



\printAffiliationsAndNotice{}  

\begin{abstract}
Capturing the occurrence dynamics is crucial to predicting \emph{which type} of events will happen next and \emph{when}. A common method to do this is Hawkes processes. To enhance their capacity, recurrent neural networks (RNNs) have been incorporated due to RNNs' successes in processing sequential data such as languages.
Recent evidence suggests self-attention is more competent than RNNs in dealing with languages. 
However, we are unaware of the effectiveness of self-attention in the context of Hawkes processes. 
This study attempts to fill the gap by designing a \emph{self-attentive Hawkes process} (SAHP). 
The SAHP employed self-attention to summarize influence from history events and compute the probability of the next event. 
One deficit of the conventional self-attention is that position embeddings only considered order numbers in a sequence, which ignored time intervals between temporal events. 
To overcome this deficit, we modified the conventional method by translating time intervals into phase shifts of sinusoidal functions. 
Experiments on goodness-of-fit and prediction tasks showed the improved capability of SAHP.
Furthermore, the SAHP is more interpretable than RNN-based counterparts because the learnt attention weights revealed contributions of one event type to the happening of another type.
To the best of our knowledge, this is the first work that studies the effectiveness of self-attention in Hawkes processes. 
\end{abstract}

\section{Introduction}
\label{sec:introduction}

\begin{figure*}
	\centering
	\includegraphics[width=0.6\textwidth]{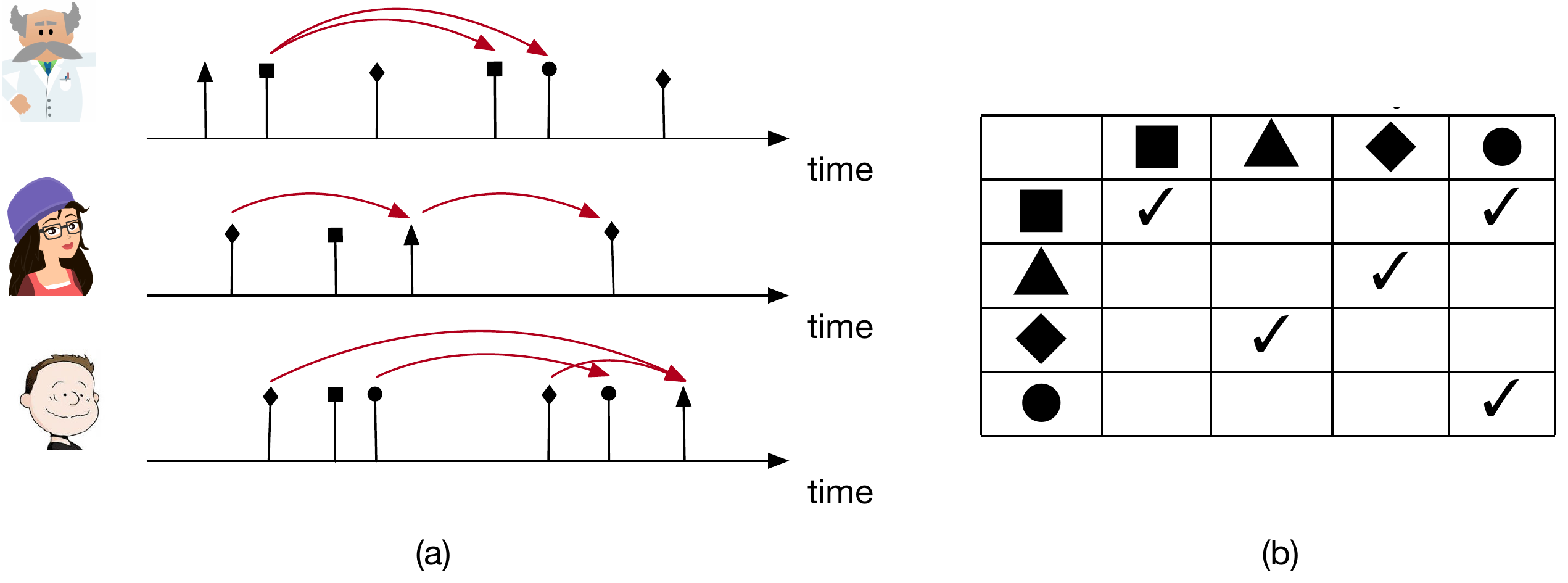}
	\vspace{-1em}
	\caption {Three users on social media platforms exert different types of actions. The filled dark symbols in (a) mean four action types while the red arrows denote actions influencing another actions. 
	A $\checkmark$ symbol in the cell $(i,j)$ in (b) indicates the influence of the column event type $j$ on the row type $i$ on future events. }
	\centering
	\label{fig:event_seq}
	\vspace{-1em}
\end{figure*}

Humans and natural phenomena often generate a large amount of irregular and asynchronous event sequences. 
These sequences can be, for example, 
user activities on social media platforms~\citep{farajtabar2015coevolve}, 
high-frequency financial transactions~\citep{bacry2014hawkes}, 
healthcare records~\citep{wang2016isotonic}, 
gene positions in bioinformatics~\citep{reynaud2010adaptive}, or 
earthquakes and aftershocks in geophysics~\citep{ogata1998space}.
Three characteristics make these event sequences unique, their: asynchronicity, multi-modality, and cross-correlation. 
A sequence is asynchronous when multiple events happening in the continuous time domain are sampled with unequal intervals. In contrary to discrete sequences where events have equal sampling intervals.
A sequence is multi-modal when sequences contain multiple type of events.
%
A sequence is cross-correlated when the occurrence of one type of event at a certain time can excite or inhibit the happening of future events of the same or another type.
Figure~\ref{fig:event_seq} shows four types of events and their mutual influence.
A classic problem with these sequences is to predict \emph{which} type and \emph{when} future events will 
happen.

The occurrence of asynchronous event sequences are often modeled by temporal point processes (TPPs)~\citep{cox1980point,brillinger2002point}. 
They are stochastic processes with (marked) events on the continuous time domain.
One special but significant type of TPPs is the Hawkes process. 
A considerable amount of studies have used Hawkes process as a \emph{de facto} standard tool to model event streams, including: 
topic modeling and clustering of text document~\citep{he2015hawkestopic,du2015dirichlet}, 
construction and inference on network structure~\citep{yang2013mixture,choi2015constructing,Etesami:2016:LNM:3020948.3020966}, 
personalized recommendations based on users’ temporal behavior~\citep{du2015time}, 
discovering of patterns in social interaction~\citep{guo2015bayesian,lukasik2016hawkes}, and 
learning causality~\citep{xu2016learning}. 
Hawkes processes usually model the occurrence probability of an event with a so called \emph{intensity function}.
For those events whose occurrence are influenced by history, the intensity function is specified as history-dependent.

The vanilla Hawkes processes specify a fixed and static intensity function, which limits the capability of capturing complicated dynamics. 
To improve its capability, recurrent neural networks (RNNs) have been incorporated as result of their success in dealing with sequential data such as speech and language. 
RNN-based Hawkes processes use a recurrent structure to summarize history events, either in the fashion of discrete-time~\citep{du2016recurrent,xiao2017modeling} or continuous-time ~\citep{mei2017neural}. 
This solution brings two benefits: 
(1) historical contributions are not necessarily addictive, and 
(2) allows for complex memory effects such as delay. 
However, recent developments in natural language processing (NLP) have led to an increasing interest in the self-attention mechanism. Although self-attention is empirically superior to RNNs in processing word sequences, it has yet to be researched whether self-attention is capable of processing event sequences that are asynchronous, multi-modal and cross-correlated.


In this work we investigate the usefulness of self-attention to Hawkes processes by 
proposing a \emph{Self-Attentive Hawkes Process} (SAHP).
First, we employ self-attention to measure the influence of historical events to the next event by computing its probability. 
As self-attention relies on positional embeddings to take into account the order of events, 
conventional embedding methods are based on sinusoidal functions where each position is distanced by a constant shift of phase, which if used for our sequences would ignore the actual time interval between events. 
We remedy this deficiency by proposing a time shifted position embedding method: time intervals act as phase shifts of sinusoidal functions. 
Second, we argue that the proposed SAHP model is more interpretable than the RNN-based counterparts: The learnt attention weights can reveal contributions of one event type to the happening of another.

The contributions of this paper can be summarized as follows:
\begin{itemize}
\item To the best of our knowledge, this work is the first to link self-attention to Hawkes processes. SAHP inherits improved capability of capturing complicated dynamics and is more interpretable.

\item To take inter-event time intervals into consideration, we propose a novel time shifted position embedding method that translates time intervals into phase shifts of sinusoidal functions.

\item Through extensive experiments on one synthetic dataset and four real-world datasets with different sequence lengths and different numbers of event types, we demonstrate the superiority of SAHP.
\end{itemize}

\section{Notation}
\label{sec:notation}

In this section we introduce the notation used throughout the paper.

\begin{tabular}{ll}
\toprule
Symbol & Description \\ 
\hline
$\mathcal{U}$ & a set of event types. \\
$\mathcal{S}$ & an event sequence. \\
$t$ & the time of an event. \\
$u,v$ & the type of an event. \\
$i,j$ & the order number of an event in a sequence. \\
$N_u(t)$ & the counting process for the events of type $u$.\\
$\mathcal{H}_t$ & the set of events that happened before time $t$. \\
$\lambda^*(t)$ & the conditional intensity function. \\
$p^*(t)$ & the conditional probability density function. \\
$F^*(t)$ & the cumulative distribution function. \\ 
\bottomrule
\end{tabular}

\section{Background}
\label{sec:background}

\subsection{Temporal Point Processes and Hawkes Process}

A temporal point process (TPP) is a stochastic process whose realization is a list of discrete events at time $t \in \mathbb{R}^+$ ~\citep{cox1980point,daley2007introduction}. 
A marked TPP allocates a type (a.k.a.~mark) $u$ to each event.
TPPs can be equivalently represented as a counting process $N(t)$, which records the number of events that have happened till time $t$. 
A multivariate TPP describes the temporal evolution of multiple event types $\mathcal{U}$. 

We indicate with $\mathcal{S}=\{(v_i,t_i)\}_{i=1}^{L}$ an event sequence, where the tuple $(v_i, t_i)$ is the $i$-th event of the sequence $\mathcal{S}$, $v_i \in \mathcal{U}$ is the event type, and $t_i$ is the timestamp of the $i$-th event. 
We indicate with $\mathcal{H}_t:=\left\{\left(v', t'\right) | t'<t,v'\in\mathcal{U}\right\}$ the historical sequence of events that happened before $t$.

Given an infinitesimal time window $[t, t+\mathrm{d}t)$, the intensity function of a TPP is defined as the probability of the occurrence of an event $(v', t')$ in $[t,t+\mathrm{d}t)$ conditioned on the history of events $\mathcal{H}_t$:
\begin{align}
\lambda^*(t)\,\mathrm{d}t :&= 
P((v', t') : t' \in [t,t+\mathrm{d}t)|\mathcal{H}_t)\nonumber \\
& = \mathbf{E}(\mathrm{d}N(t)|\mathcal{H}_t),
\end{align}
where $\mathbf{E}(\mathrm{d}N(t)|\mathcal{H}_t)$ denotes the expected number of events in $[t,t+\mathrm{d}t)$ based on the to the history $\mathcal{H}_t$. Without loss of generality, we assume that two events do not happen simultaneously, i.e., $\mathrm{d} N(t)\in\{0,1\}$.

Based on the intensity function, it is straightforward to derive the probability density function $p^*(t)$ and the cumulative density function $F^*(t)$~\citep{rasmussen2018lecture}: 
\begin{equation}
    p^*(t)=\lambda^{*}(t) \exp \left(-\int_{t_{i-1}}^{t} \lambda^{*}(\tau) \mathrm{d} \tau\right),
\end{equation}
\begin{equation}
    F^*(t)=1-\exp \left(-\int_{t_{i-1}}^{t} \lambda^{*}(\tau) \mathrm{d} \tau\right).
\end{equation}


%
An Hawkes process~\citep{hawkes1971hausdorff} models the 
self-excitation of events of the same type and the mutual excitation of different event types, in an additive way. 
Hence, the definition of the intensity function is given as:
\begin{equation}\label{eq:hawkes}
    \lambda^*(t) = \mu + \sum_{(v',t') \in \mathcal{H}_t} \phi(t-t'),
\end{equation}
where $\mu \geq 0$ (a.k.a. \emph{base intensity}) is an exogenous component of the intensity function independent of the history, 
while $\phi(t) > 0$ is an endogenous component dependent on the history. 
Besides, $\phi(t)$ is a triggering kernel containing the peer influence of different event types. 
To highlight the peer influence represented by $\phi(t)$, we write $\phi_{u,v}(t)$, which captures the impact of a historical type-$v$ event on a subsequent type-$u$ event~\citep{farajtabar2014shaping}. 
In this example, the occurrence of a past type-$v$ event increases the intensity function $\phi_{u,v}(t-t')$ for $0<t'<t$. 

Most commonly $\phi_{u,v}(t)$ is parameterized as $\phi_{u,v}(t)=\alpha_{u,v} \cdot \kappa(t) \cdot \mathbbm{1}_{t>0}$~\citep{zhou2013learningsocial,xu2016learning}. The \emph{excitation} parameter $\alpha_{u,v}$ quantifies the initial influence of the type-$v$ event on the intensity of the type-$u$ event. 
The \emph{kick} function $\kappa(t)$ characterizes the time-decaying influence. 
Typically, $\kappa(t)$ is chosen to be exponential, i.e., $\kappa(t)= \exp({-\gamma t})$, where $\gamma$ is the \emph{decaying} parameter controlling the intensity decaying speed. 



To learn the parameters of Hawkes processes, it is common to use Maximum Likelihood Estimation (MLE). 
Other advanced and more complex 
adversarial learning~\citep{xiao2017wasserstein} and
reinforcement learning~\citep{li2018learning} methods have been proposed, however we use MLE for its simplicity. 
In experiments, we use the same optimization method for our model and all baselines as done in their original papers. 
To apply MLE, a loss function is derived based on the negative log-likelihood. Details of derivation can be found in appendix.
The likelihood of a multivariate Hawkes process over a time interval $[0,T]$ is given by:
\begin{equation}
\label{eq:obj}
    \mathcal{L} (\lambda) = \sum^{L}_{i=1} \log \lambda_{v_{i}} (t_i) - \int_0^T \lambda(\tau)d\tau,
    \vspace{-0.1em}
\end{equation}
where the first term is the sum of the log-intensity functions of past events, and the second term corresponds to the log-likelihood of infinitely many non-events. 
Intuitively, the probability that there is no event of any type in the infinitesimally time interval $[t,t+dt)$ is equal to $1-\lambda(t) dt$, the log of which is $-\lambda(t)dt$.

\subsection{Attention and Self-Attention}

\paragraph{Attention.} 
The attention mechanism enables machine learning models to focus on a subset of the input sequence~\citep{walther2004usefulness,DBLP:journals/corr/BahdanauCB14}. 
In Seq2Seq models with the attention mechanism the input sequence, in the encoder, is represented with a sequence of key vectors $K$ and value vectors $V$, $(K,V)=[(\boldsymbol{k}_1,\boldsymbol{v}_1), (\boldsymbol{k}_2,\boldsymbol{v}_2), \dots, (\boldsymbol{k}_N,\boldsymbol{v}_N)]$. 
While, the decoder side of the Seq2Seq model uses query vectors, $Q=[\boldsymbol{q}_1, \boldsymbol{q}_2, \dots, \boldsymbol{q}_M]$.  These query vectors are used to find which part of the input sequence is more contributory~\citep{DBLP:conf/nips/VaswaniSPUJGKP17}.
Given these two sequences of vectors $(K,V)$ and $Q$, the attention mechanism computes a prediction sequence $O = [\boldsymbol{o}_1, \boldsymbol{o}_2, \dots, \boldsymbol{o}_M]$ as follows:
\begin{equation}
     \boldsymbol{o}_m = \left( \sum_{n}f(\boldsymbol{q}_m,\boldsymbol{k}_n) g(\boldsymbol{v}_n) \right) / \sum_{n}f(\boldsymbol{q}_m,\boldsymbol{k}_n),
     \vspace{-0.5em}
\end{equation}
where $m \in \{1,\dots,M\}$, $n \in \{1,\dots,N\}$, $\boldsymbol{q}_m \in \mathbb{R}^d$, $\boldsymbol{k}_n \in \mathbb{R}^d$, $\boldsymbol{v}_n \in \mathbb{R}^p$, $g(\boldsymbol{v}_n) \in \mathbb{R}^q$ and $\boldsymbol{o}_m \in \mathbb{R}^q$. 
The similarity function $f(\boldsymbol{q}_m,\boldsymbol{k}_n)$ characterizes the relation between $\boldsymbol{q}_m$ and $\boldsymbol{k}_n$, 
whose common form is composed of: 
an embedded Gaussian, 
an inner-product, and 
a concatenation~\citep{wang2018non}. 
The function $g(\boldsymbol{v}_n)$ is a linear transformation specified as $g(\boldsymbol{v}_n):=\boldsymbol{v}_n W_v$, where $W_v \in \mathbb{R}^{p\times q}$ is a weight matrix.

\paragraph{Self-attention.}
Self-attention is a special case of the attention mechanism~\citep{DBLP:conf/nips/VaswaniSPUJGKP17}, where the query vectors $Q$, like $(K,V)$, 
are from the encoder side. 
Self-attention is a method of encoding sequences of input tokens by relating these tokens to each other based on a pairwise similarity function $f(\cdot,\cdot)$. 
It measures the dependency between each pair of tokens from the same input sequence.
To encode position information of tokens, position embeddings are calculated based on order numbers in a sequence. Consequently, self-attention encodes both token similarity and position information.

Self-attention is very expressive and flexible for both long-term and local dependencies, which used to be modeled by recurrent neural networks (RNNs) and convolutional neural networks (CNNs)~\citep{DBLP:conf/nips/VaswaniSPUJGKP17}. 
Moreover, the self-attention mechanism has fewer parameters and faster convergence than RNNs. 
Recently, a variety of Natural Language Processing (NLP) tasks have experienced large improvements thanks to self-attention~\citep{DBLP:conf/nips/VaswaniSPUJGKP17,devlin2018bert}.

\section{Self-Attentive Hawkes Process}
\label{sec:sah}
\begin{figure*}[!ht]
	\centering
	\includegraphics[width=0.75\textwidth]{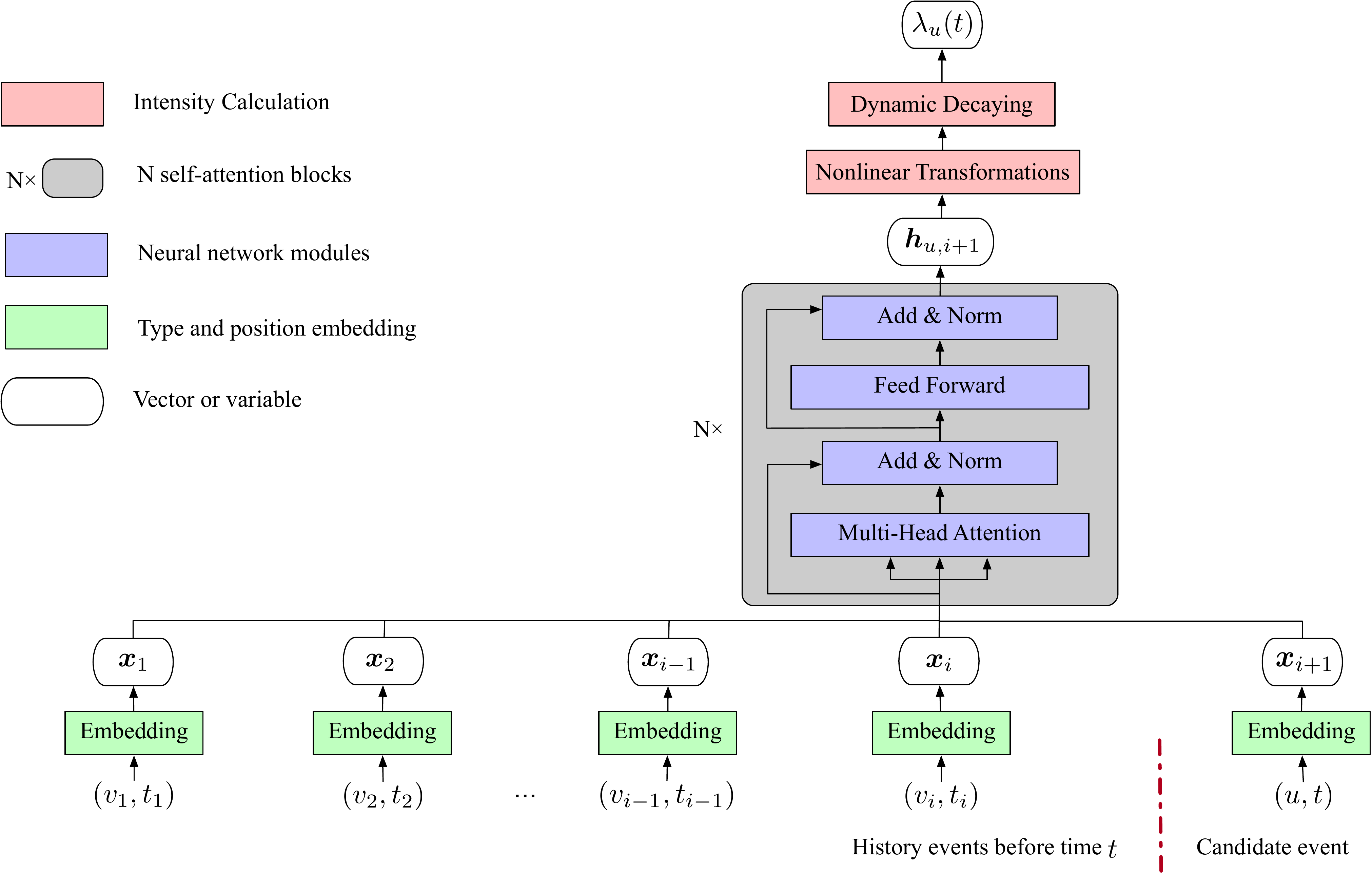}
	\vspace{-0.5em}
	\caption{An event stream and the SAHP for one event type ($u$). 
	The intensity function ($\lambda_u(t)$) is determined by a sequence of past events via the SAHP. 
	The length of each temporal evolution arrow represents the time interval between subsequent events.}
	\centering
	\label{fig:framework}
	\vspace{-1em}
\end{figure*}

In this section, we describe how to adapt the self-attention mechanism to Hawkes processes as Figure~\ref{fig:framework} shows.

%
\paragraph{Event type embedding.}
The input sequence is made up of events. To obtain a unique dense embedding for each event type, we use a linear embedding layer,
\begin{equation}
    \boldsymbol{tp}_v = \boldsymbol{e}_v W_E,
\end{equation}
where 
$\boldsymbol{tp}_v$ is the type-$v$ embedding, 
$\boldsymbol{e}_v$ is a one-hot vector of the type-$v$ and 
$W_E$ is the embedding matrix.

\paragraph{Time shifted positional encoding.}

\begin{figure}[!ht]
	\centering
	\includegraphics[width=0.5\textwidth]{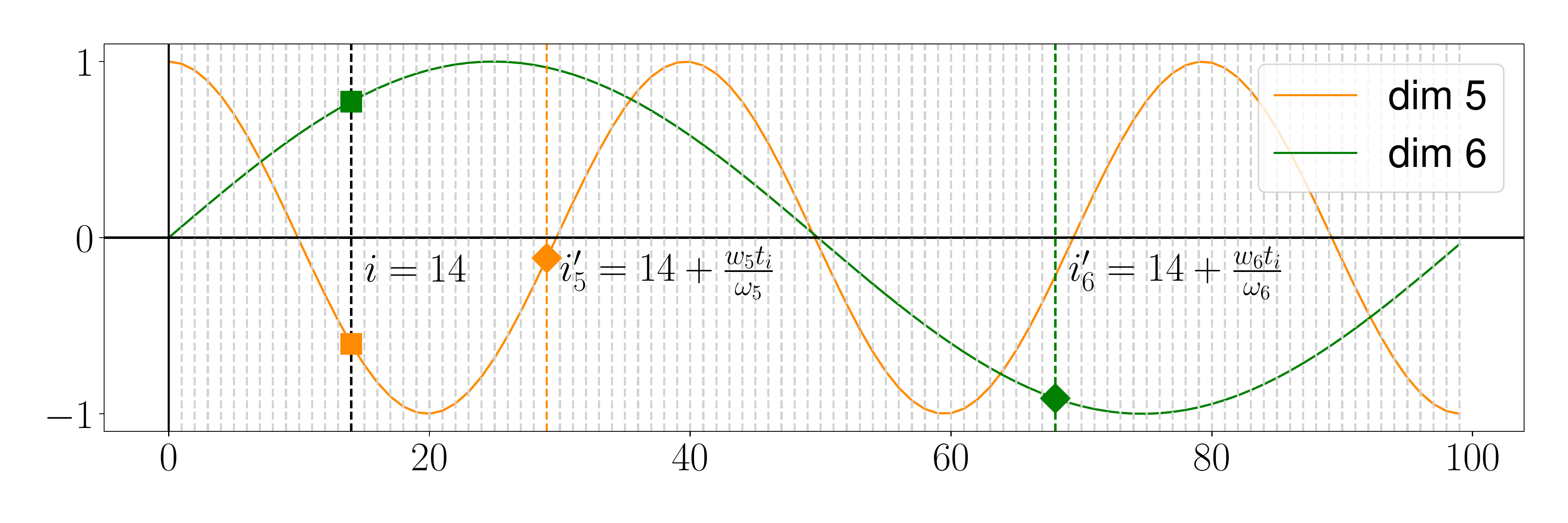}
	\vspace{-2.5em}
	\caption{The time shifted position embedding of an event with $i=14$ in a sequence. Squares and diamonds denote conventional and new embedding values with time shifts.}
	\centering
	\label{fig:position_emb}
	\vspace{-3em}
\end{figure}

Self-attention utilizes positional encoding to inject order information to a sequence. To take into account time intervals of subsequent events, we modify the conventional position encoding. 
For an event $(v_i, t_i)$, the positional encoding is defined as a $K$-dimensional vector such that the $k$-th dimension of the position embedding is calculated as:
\begin{align}
    pe^{k}_{(v_i, t_i)} = \sin{(\omega_k \times i + w_k \times t_i)},
\end{align}
where $i$ is the absolute position of an event in a sequence, and $\omega_k$ is the angle frequency of the $k$-th dimension, which is pre-defined and will not be changed. While $w_k$ is a scaling parameter that converts the timestamp $t_i$ to a phase shift in the $k$-th dimension.
Multiple sinusoidal functions with different $\omega_k$ and $w_k$ are used to generate the multiple position values, the concatenation of which is the new positional encoding.
Even and odd dimensions of $\boldsymbol{pe}$ are generated from $\sin{}$ and $\cos{}$ respectively. 

Figure~\ref{fig:position_emb} shows how conventional and the new positional encodings work. 
Suppose an event $(v_i, t_i)$ is at the $i=14$ position of a sequence. 
Conventional methods calculate the values of sinusoidal functions at the $i=14$ position as the position value of this event. Our encoding modifies this by shifting the original position $i$ to a new position $i^\prime_k=i+\frac{w_k t_i}{\omega_k}$, where $k$ denotes the embedding dimension. This is equivalent to interpolating the time domain and to produce shorter equal-length time periods. Positions in a sequence are thus shifted by the time $t_i$. The length of time periods is decided by $\frac{w_k}{\omega_k}$. 
Since $w_k$ and $\omega_k$ are dimension-specific, the shift in one dimension can be different from others. 

\paragraph{Historical hidden vector.}
As an event consists of its type and timestamp, we add the positional encoding to the event type embedding in order to obtain the representation of the event $(v_i,t_i)$:
\begin{equation}
    \boldsymbol{x}_{i} = \boldsymbol{tp}_v + \boldsymbol{pe}_{(v_i, t_i)}.
\end{equation}

\paragraph{Self-Attention.}
Given a series of historical events until $t_i$, 
to compute the intensity of the type-$u$ at the timestamp $t$, we need to consider the influence of all types of events before it. To do this, we compute the pairwise influence of one previous event to the next event by employing self-attention. This generates a hidden vector that summarizes the influence of all previous events:
\begin{align}
  \boldsymbol{h}_{u,i+1} =
  \left( \sum_{j = 1}^{i} f(\boldsymbol{x}_{i+1},\boldsymbol{x}_{j}) g(\boldsymbol{x}_{j}) \right) / 
  \sum_{j = 1}^{i} f(\boldsymbol{x}_{i+1},\boldsymbol{x}_{j}),
  \vspace{-1em}
\end{align}
where $\boldsymbol{x}_{i+1}$ is like query ($\boldsymbol{q}$) in the attention terminology, 
$\boldsymbol{x}_{j}$ is the key ($\boldsymbol{k}$) and $g(\boldsymbol{x}_{j})$ is the value ($\boldsymbol{v}$). The function $g(\cdot)$ is a linear transformation while the similarity function $f(\cdot, \cdot)$ is specified as an embedded Gaussian:
\begin{equation}
f(\boldsymbol{x}_{i+1},\boldsymbol{x}_{j}) = 
\exp\left(\boldsymbol{x}_{i+1} \boldsymbol{x}_{j}^T\right).
\end{equation}

%
%
%
%

The temporal information is provided to the model during training by preventing the model to learn about future events via masking. 
We implement this in the attention mechanism by masking out all values in the input sequence that correspond to future events. 
Hence, the intensity of one event is obtained only based on its history.

\paragraph{Intensity function.}
Since the intensity function of Hawkes processes is history-dependent, we compute three parameters of the the intensity function based on the history hidden vector $\boldsymbol{h}_{u,i+1}$ via the following three non-linear transformations:
\begin{align}
    \mu_{u, i+1} &=  \gelu\left(\boldsymbol{h}_{u,i+1} W_\mu \right),\quad\\
    \eta_{u,i+1} &=  \gelu\left(\boldsymbol{h}_{u,i+1} W_\eta \right),\quad \\
    \gamma_{u,i+1} &= \softplus\left(\boldsymbol{h}_{u,i+1} W_\gamma \right).
\end{align}

The function $\gelu$ represents the Gaussian Error Linear Unit for nonlinear activations. We use this activation function because this has been empirically proved to be superior to other activation functions for self-attention~\cite{hendrycks2016gaussian}.
$\softplus$ is used for the decaying parameter since $\gamma$ needs to be constrained to strictly positive values. 

Finally, we express the intensity function as follows:
\begin{align}
\label{eq:int_func}
    \lambda_{u}(t)  = &\softplus (\mu_{u,i+1} + \nonumber \\
    &(\eta_{u,i+1}-\mu_{u,i+1}) \exp(-\gamma_{u,i+1} (t-t_i))), 
\end{align}   
$\text{for}\ t \in (t_{i},t_{i+1}]$,
where the $\softplus$ is employed to constrain the intensity function to be positive.
The starting intensity at $t=t_i$ is $\eta_{u,i+1}$. When $t$ increases from $t_i$, the intensity decays exponentially. As $t \rightarrow \infty$, the intensity converges to $\mu_{u,i+1}$. The decaying speed is decided by $(\eta_{u,i+1}-\mu_{u,i+1})$ that can be both positive and negative. This enables us to capture both excitation and inhibition effects. With inhibition we mean the effect when past events reduce the likelihood of future events~\citep{mei2017neural}.

\section{Experiments}
\label{sec:experiments}
\begin{table*}[!ht]
\centering
\vspace{-0.5em}
\caption{Statistics of the used datasets.}
\vspace{0.5em}
\begin{threeparttable}
\begin{tabular}{@{}lrrrrrrr@{}}
\hline\hline
Dataset       & \multicolumn{1}{c}{\# of Types} & \multicolumn{3}{c}{Sequence Length}                                          & \multicolumn{3}{c}{\# of Sequences}                                                \\ \cmidrule(l){3-8} 
              & \multicolumn{1}{l}{}              & \multicolumn{1}{c}{Min} & \multicolumn{1}{c}{Mean} & \multicolumn{1}{c}{Max} & \multicolumn{1}{c}{Train} & \multicolumn{1}{c}{Validation} & \multicolumn{1}{c}{Test} \\ \midrule
Synthetic      & 2                                 & 68                      & 132                      & 269                     & 3,200                 & 400                        & 400                  \\
RT      & 3                                 & 50                      & 109                      & 264                     & 20,000                 & 2,000                        & 2,000                  \\
SOF & 22                                & 41                      & 72                       & 736                     & 4,777 & 530 & 1,326                   \\
MMC     & 75                             & 2                      & 4                        & 33                      & 527 & 58 & 65                   \\
\hline\hline
\end{tabular}

\end{threeparttable}
\label{tbl:dataset}
\vspace{-1em}
\end{table*}

To compare our method with the state-of-the-art, we conduct experiments on one synthetic dataset and four real-world datasets.
The datasets have been purposefully chosen in order to span over various properties, i.e., the number of event type ranges from $2$ to $75$ and the average sequence length ranges from $4$ to $132$. As usual, sequences from the same dataset are assumed to be drawn independently from the same process. 
Each dataset is split into a training set, a validation set and a testing set. The validation set is used to tune the hyper-parameters while the testing set is used to measure the model performance. Details about the datasets can be found in Table~\ref{tbl:dataset} and Appendix. 
These datasets are all available at the following weblink\footnote{\url{https://drive.google.com/drive/folders/0BwqmV0EcoUc8UklIR1BKV25YR1U}}.


\subsection{Synthetic Dataset}

\begin{figure}[!ht]
	\centering
	\vspace{-1em}
	\includegraphics[width=0.4\textwidth]{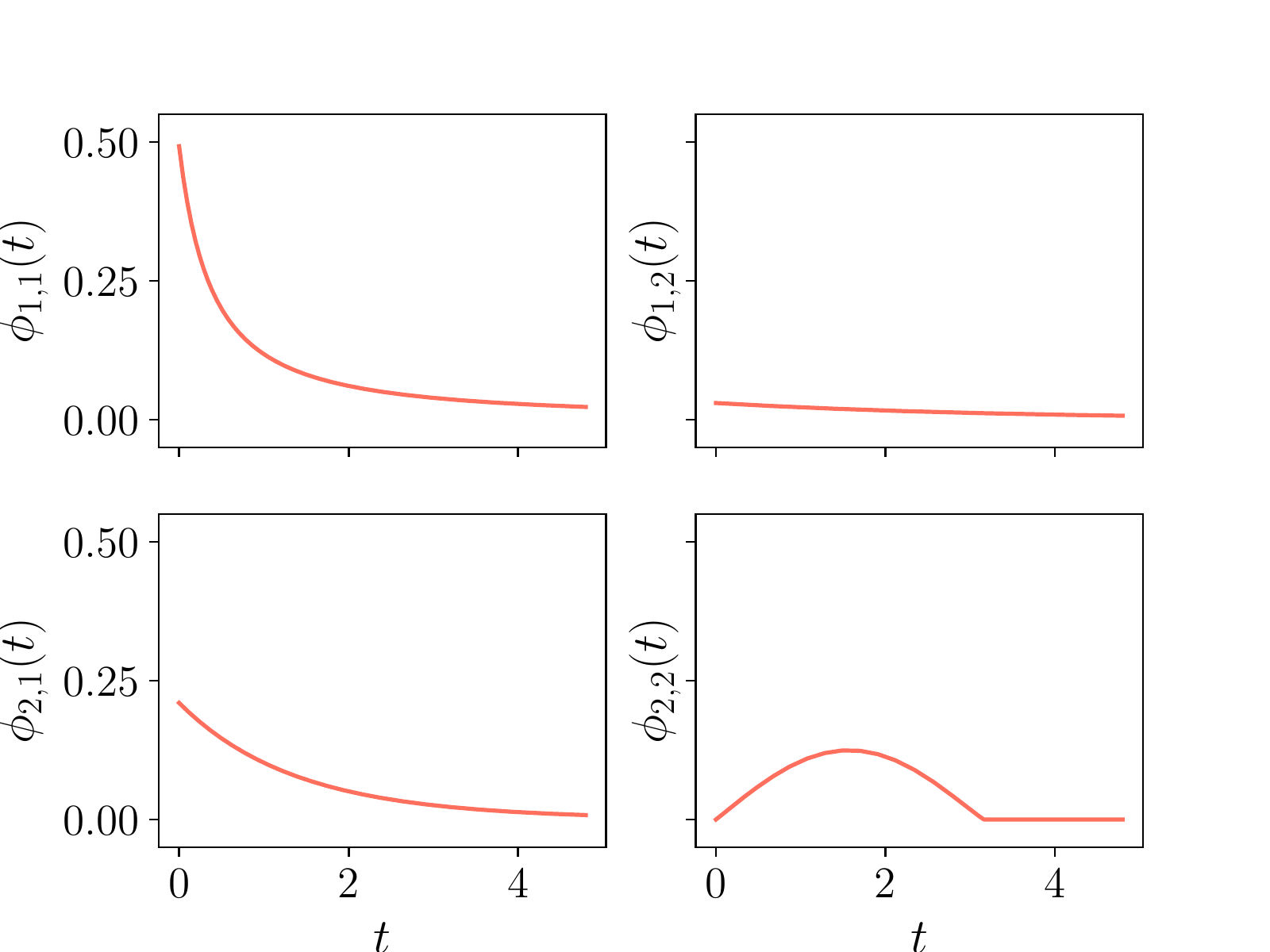}
	\vspace{-1em}
	\caption{The four triggering kernels of the synthetic dataset with 2 event types.
	}
	\centering
	\label{fig:kernels}
	\vspace{-1em}
\end{figure}

We generate a synthetic dataset using the open-source Python library \emph{tick}\footnote{\url{https://github.com/X-DataInitiative/tick}}. A two-dimensional Hawkes process is generated with base intensities $\mu_1=0.1$ and $\mu_2=0.2$. The triggering kernels consist of a power law kernel, an exponential kernel, a sum of two exponential kernels, and a sine kernel:
\begin{align}
    \phi_{1,1}(t) &= 0.2\times (0.5+t)^{-1.3}\\
    \phi_{1,2}(t) &= 0.03\times \exp(-0.3t)\\
    \phi_{2,1}(t) &=  0.05\times \exp(-0.2t)+0.16\times \exp(-0.8t)\\
    \phi_{2,2}(t) &= \max (0, \sin(t)/8 ) \quad \text{for}\,\,0\leq t\leq4
\end{align}
%
In Figure~\ref{fig:kernels} we show the four triggering kernels of the 2-dimensional Hawkes processes.
The simulated intensities of each dimension is shown in appendix.

\subsection{Training Details}
\label{subsec:training}
We implement the multi-head attention. This allows the model to jointly attend  information from different representation subspaces~\citep{DBLP:conf/nips/VaswaniSPUJGKP17}. 
The number of heads is a hyper-parameter. 
We explore this hyper-parameter in the set $\{1, 2, 4, 8, 16\}$. 
Another hyper-parameter is the number of attention layers. We explore this hyper-parameter in the set $\{2,3,4,5,6\}$. 
We adapt the Adam as the basic optimizer and develop a warm-up stage for the learning rate whose initialization is set to $1e{-4}$.
To mitigate overfitting we apply dropout with rate set to 0.1. 
Early stopping is used when the validation loss does not decrease more than $1e{-3}$.

\subsection{Baselines}

%
    \paragraph{Hawkes Processes (HP).} This is the most conventional Hawkes process statistical model which intensity is described in Eq.~\ref{eq:hawkes}. It uses an exponential kernel;
    \paragraph{Recurrent Marked Temporal Point Processes (RMTPP).}~This method \citep{du2016recurrent} uses RNN to learn a representation of influences from  past events, and time intervals are encoded as explicit inputs;
    \paragraph{Continuous Time LSTM (CTLSTM).} \citet{mei2017neural} use a continuous-time LSTM, which includes intensity decay and eliminate the need to encode event intervals as numerical inputs of the LSTM.
    \paragraph{Fully Neural Network (FullyNN).}\citet{omi2019fully} propose to model the cumulative distribution function with a feed-forward neural network.
    \paragraph{Log Normal Mixture (LogNormMix).}\citet{shchur2019intensity} suggest to model the conditional probability density distribution by a log-normal mixture model. 
%

\section{Results and Discussion}
\label{sec:results_and_discussion}
\begin{figure*}[!t]
\vspace{0.5em}
\centering
\subfigure[HP]{
\includegraphics[width=0.13\textwidth]{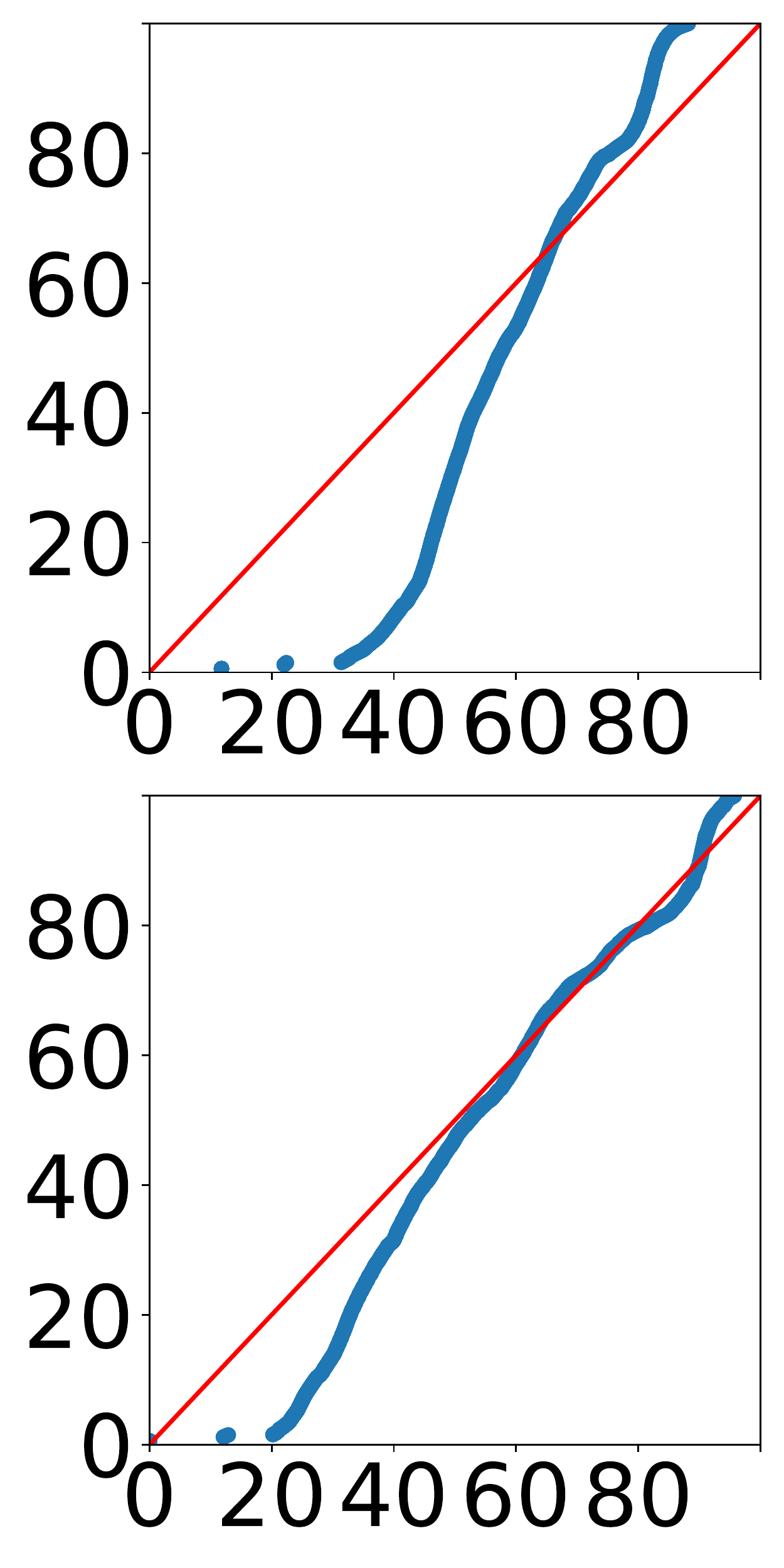}}
\subfigure[RMTPP]{
\includegraphics[width=0.13\textwidth]{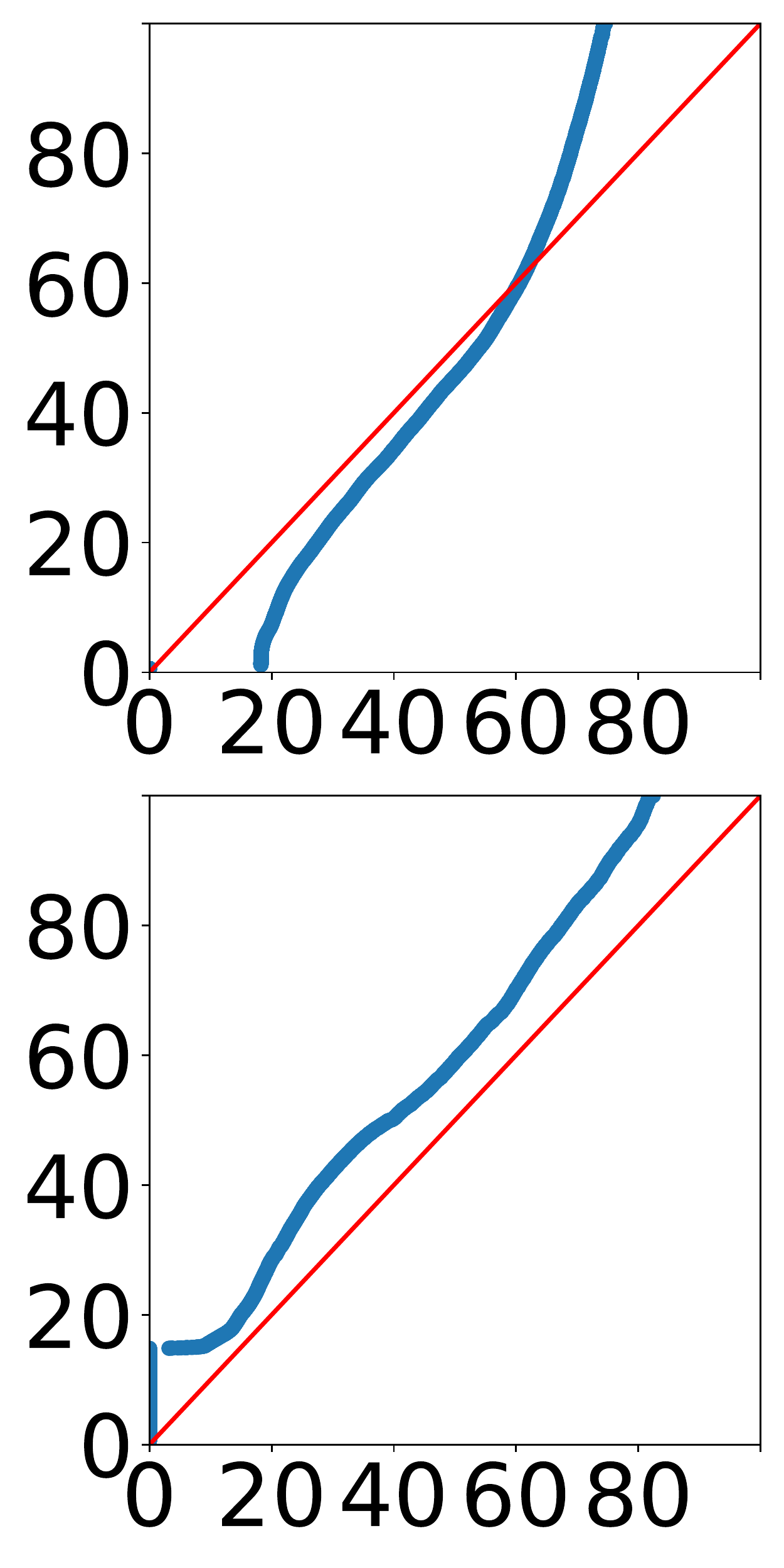}}
\subfigure[CTLSTM]{
\includegraphics[width=0.13\textwidth]{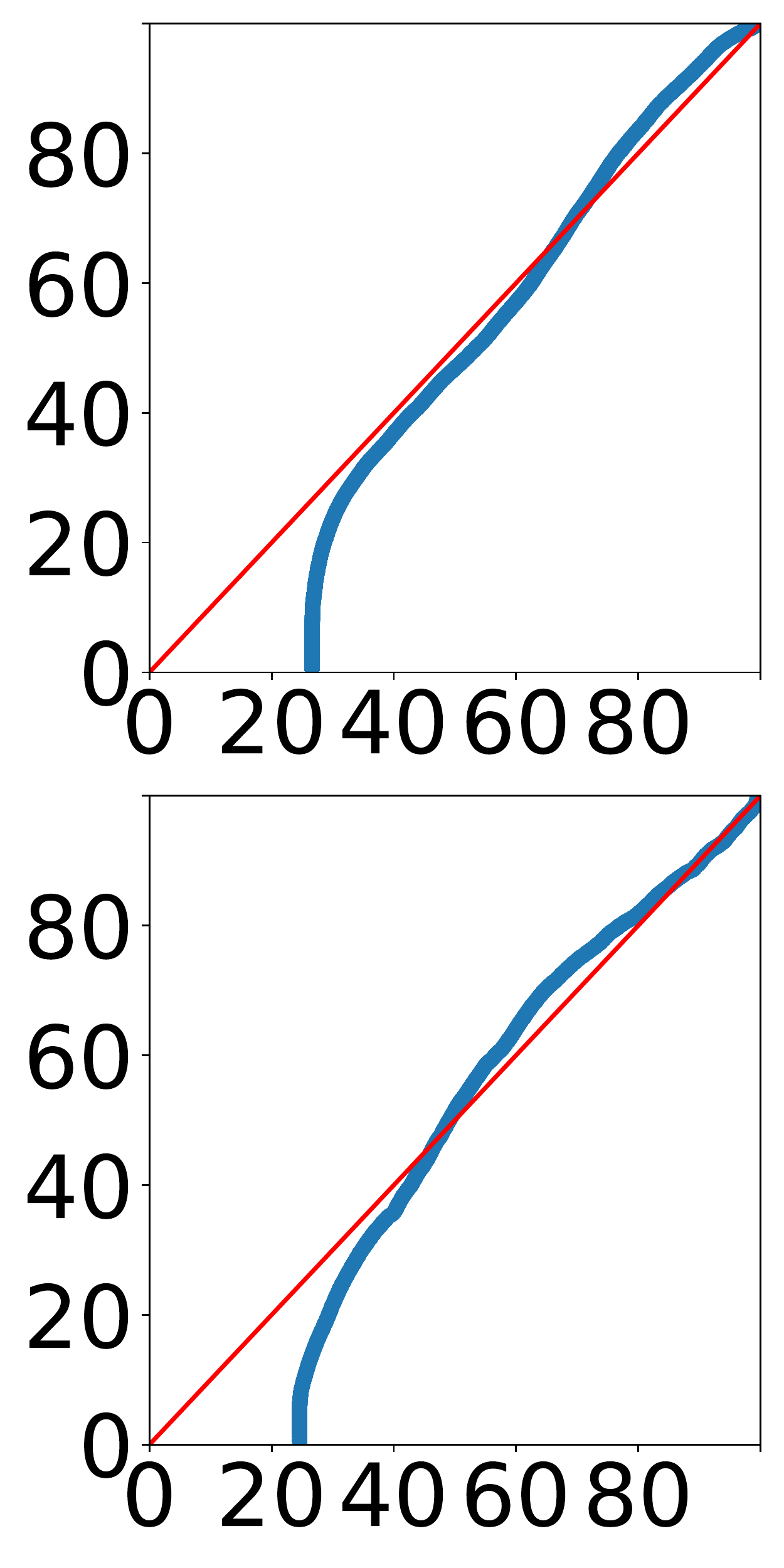}}
\subfigure[FullyNN]{
\includegraphics[width=0.13\textwidth]{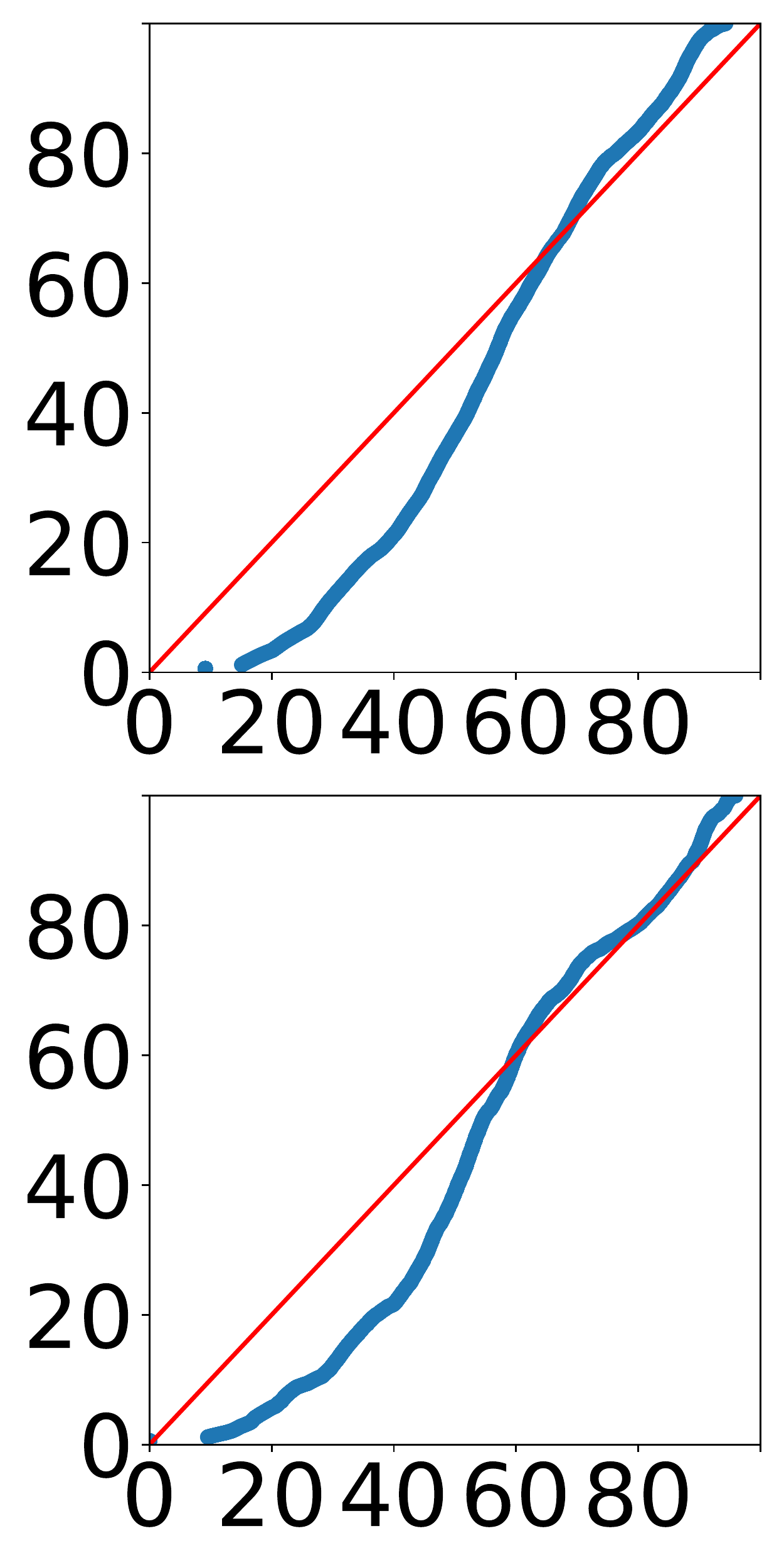}}
\subfigure[LogNormMix]{
\includegraphics[width=0.13\textwidth]{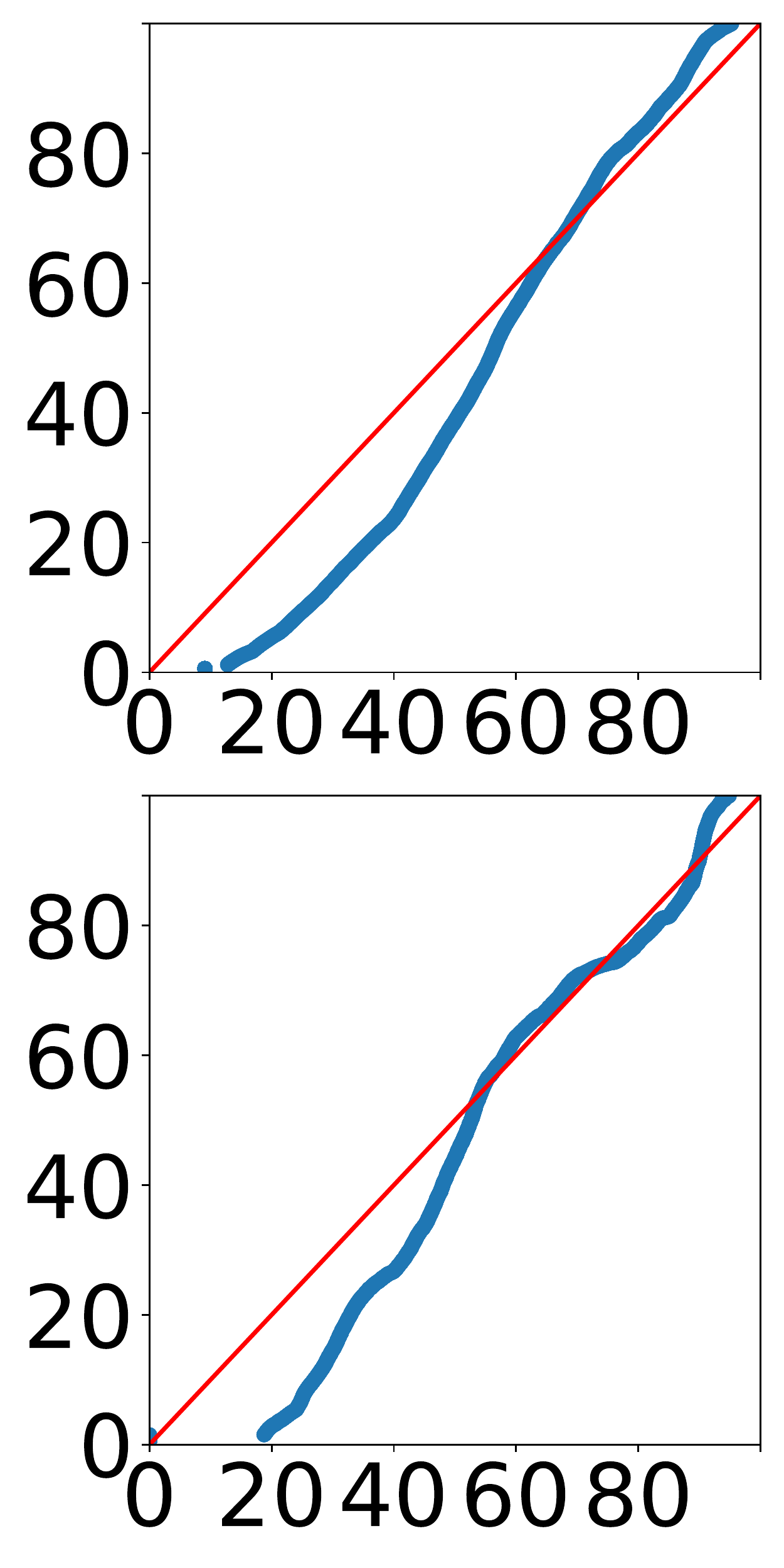}}
\subfigure[SAHP]{
\includegraphics[width=0.13\textwidth]{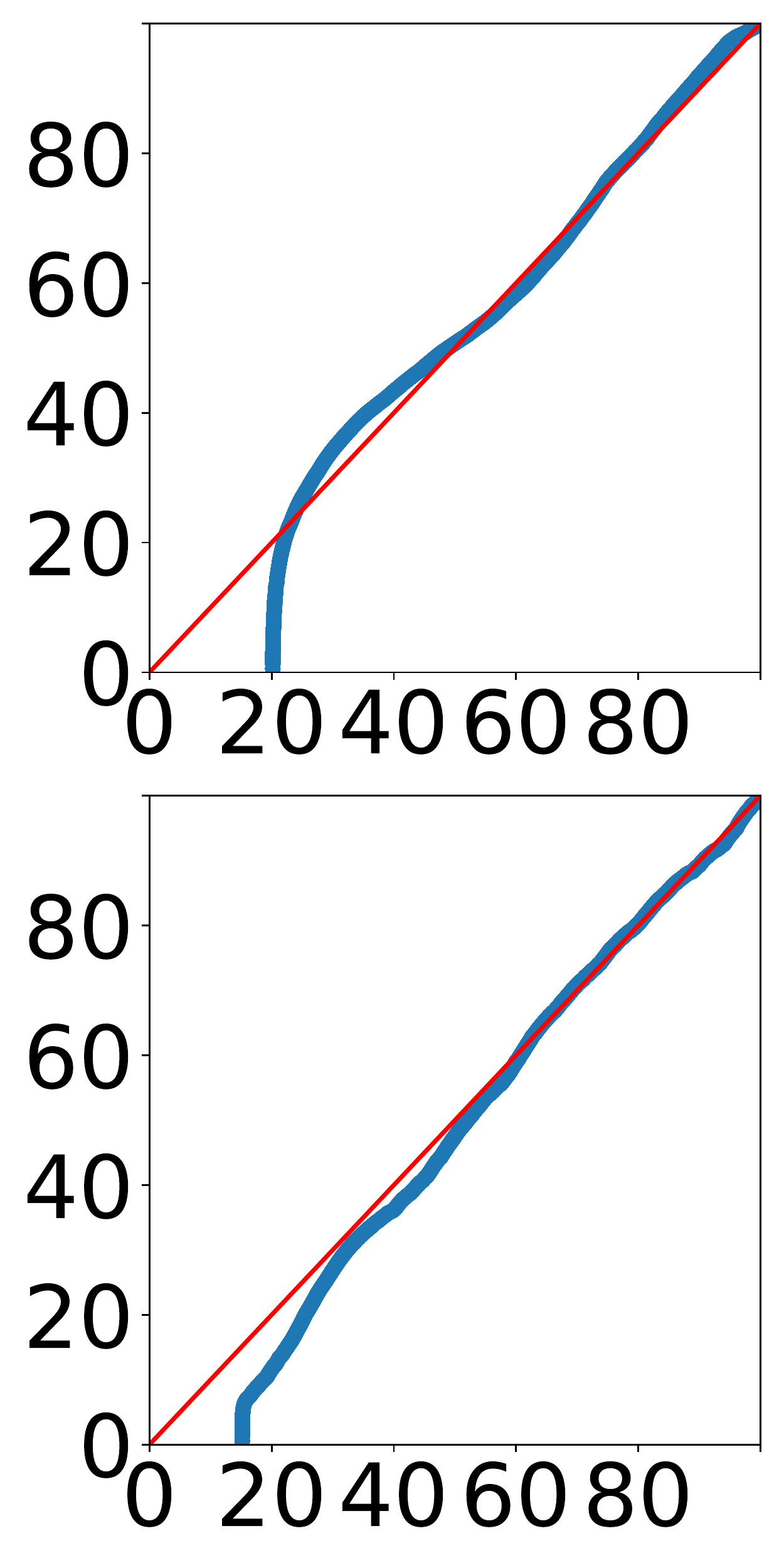}}
\vspace{-1em}
\caption{QQ-plot of true VS estimated intensities of types. The x-axis and the y-axis represent the quantiles of the true and estimated intensities. For each model the top figure is for the type-1 events while the bottom figure is for the type-2 events.}
\label{fig:qqplot}
\vspace{-1em}
\end{figure*}

For a fair comparison, we tried different hyper-parameter configurations for baselines and our model, and selected the configuration with the best validation performance.
The software used to run these experiments is 
is available at the following web-link: anonymous.

\paragraph{Goodness of fit on the synthetic dataset.}
In order to conduct a goodness-of-fit evaluation, we used the synthetic dataset where the true intensity is known, and compared the estimated intensity against the true intensity. We chose the QQ-plot to visualize how well the proposed SAHP is able to approximate the true intensity. Figure~\ref{fig:qqplot} shows the QQ-plots of intensity estimated by the five baselines and SAHP. 

From this figure, we observe that the intensity estimated by SAHP produces the most similar distribution to the true one, which indicates that SAHP is able to best capture the underlying complicated dynamics of the synthetic dataset. Moreover, by comparing the upper and the lower sub-figures in one column, all models obtain slightly better approximations to the intensity of the second event type.

\paragraph{Sequence modelling.} 
Besides evaluation of goodness-of-fit, we further compare the ability of the methods to model an event sequence. 
As done in previous works~\citep{mei2017neural, shchur2019intensity}, negative log-likelihood (NLL) was selected as the evaluation metric.
The lower the NLL is, the more capable a model is to model a specific event sequence. 

In Table~\ref{tab:performance} we report the per-event NLL of these models on each test set. 
According to Table~\ref{tab:performance}, our method significantly outperforms the baselines in all datasets.
As expected, the conventional HP method is the worst in modeling an event sequence in all datasets.
RMTPP and CTLSTM have very similar performance except on the Retweet dataset, where CTLSTM achieves a lower NLL than RMTPP. 

\begin{table}[!ht]
\centering
\vspace{-1em}
\caption{Negative log-likelihood per event on the four test sets.}
\label{tab:performance}
\begin{tabular}{l|rrrr}
\hline\hline
Dataset & \multicolumn{1}{c}{Synthetic} & \multicolumn{1}{c}{RT} & \multicolumn{1}{c}{SOF} & \multicolumn{1}{c}{MMC} \\ \hline
HP & 2.12 & 9.84 & 3.21 & 1.81  \\
RMTPP & 1.85 & 7.43 & 2.44 & 1.33 \\
CTLMST & 1.83 & 6.95 & 2.38 & 1.36  \\
FullyNN & 1.55 & 6.23 & 2.21 & 1.03 \\
LogNormMix & 1.43 & 5.32 & 2.01 & 0.78  \\
SAHP & \textbf{1.35} & \textbf{4.56} & \textbf{1.86} & \textbf{0.52} \\ 
\hline\hline
\end{tabular}
\vspace{-1em}
\end{table}

\paragraph{Event prediction.} 
We also evaluate the ability of the methods  to predict the next event, 
including type and the , according to history. To emphasizes the importance of the time shifted positional embedding, we also compare SAHP with a version (SAHP-TSE) where the new positional encoding is replaced with the standard one as in ~\citep{DBLP:conf/nips/VaswaniSPUJGKP17}. We categorize type prediction as a multi-class classification problem. As there is class imbalance among event types, we use the macro $F_1$ as the evaluation metric. Also, since time interval prediction is assumed to be a real number, a common evaluation metric to evaluate these cases is to use the Root Mean Square Error (RMSE). In order to eliminate the effect of the scale of time intervals, we compute the prediction error according to 
\begin{equation}
    \varepsilon_i = \frac{(\hat{t}_{i+1}-t_i)-(t_{i+1}-t_i)}{t_{i+1}-t_i},
\end{equation}
where $\hat{t}_{i+1}$ is the predicted  while $t_{i+1}$ is the ground truth, and $\hat{t}_{i+1}-t_i$ is the predicted time interval while $t_{i+1}-t_i$ is the true time interval. 
The results of type and time prediction are summarized in Table~\ref{tab:pred_type} and Table~\ref{tab:pred_time}. 

These two tables illustrate that our model outperforms the baselines in terms of $F_1$ and RMSE on all the prediction tasks. 
We also observe that SAHP demonstrates a larger margin in type prediction for $F_1$. 
FullyNN and LogNormMix are consistently better than the other baselines in time prediction, yet LogNormMix is not good at predicting event types, which confirms the previous findings~\citep{shchur2019intensity}. 
Another important finding is that the use of the time shifted positional embedding improves the performance of our method in both tasks. 

\begin{table}[!ht]
\centering
\caption{$F_1(\%)$ of event type prediction on the four test-sets.}
\label{tab:pred_type}
\begin{tabular}{l|rrrrr}
\hline\hline
Dataset & \multicolumn{1}{c}{Synthetic} & \multicolumn{1}{c}{RT} & \multicolumn{1}{c}{SOF} & \multicolumn{1}{c}{MMC} \\ \hline
HP & 33.20 & 32.43 & 2.98 & 19.32 \\
RMTPP & 40.32  & 41.22 & 5.44 & 28.76  \\
CTLMST & 43.80 & 39.21 & 4.88 & 34.00 \\
FullyNN & 45.21  &43.80  &6.34  &33.32   \\
LogNormMix &42.09  &45.25  &3.23  &32.86 \\
SAHP-TSE & 57.93 & 53.24 & 24.05 & 34.23 \\
SAHP & \textbf{58.50} & \textbf{53.92} & \textbf{24.12} & \textbf{36.90} \\ 
\hline\hline
\end{tabular}
\end{table}

\begin{table}[!ht]
\centering
\caption{RMSE of event  prediction on the four test sets.}
\label{tab:pred_time}
\begin{tabular}{l|rrrrr}
\hline\hline
Dataset & \multicolumn{1}{c}{Synthetic} & \multicolumn{1}{c}{RT} & \multicolumn{1}{c}{SOF} & \multicolumn{1}{c}{MMC}\\ \hline
HP    & 42.80 & 1293.32 & 221.82 & 7.68  \\
RMTPP & 37.07 & 1276.41 & 207.79 & 6.83  \\
CTLMST &35.08 & 1255.05 & 194.87 & 6.49 \\
FullyNN & 33.34 & 1104.41 & 173.92 & 5.43  \\
LogNormMix & 32.64 & 1090.45 & 154.13 & 4.12  \\
SAHP-TSE & 33.32 & 1102.34 & 143.54 & 4.03 \\
SAHP & \textbf{31.16} & \textbf{1055.05} & \textbf{133.61} & \textbf{3.89} \\ 
\hline\hline
\end{tabular}
\end{table}

\begin{figure}[h]
	\centering
	\includegraphics[width=0.5\textwidth]{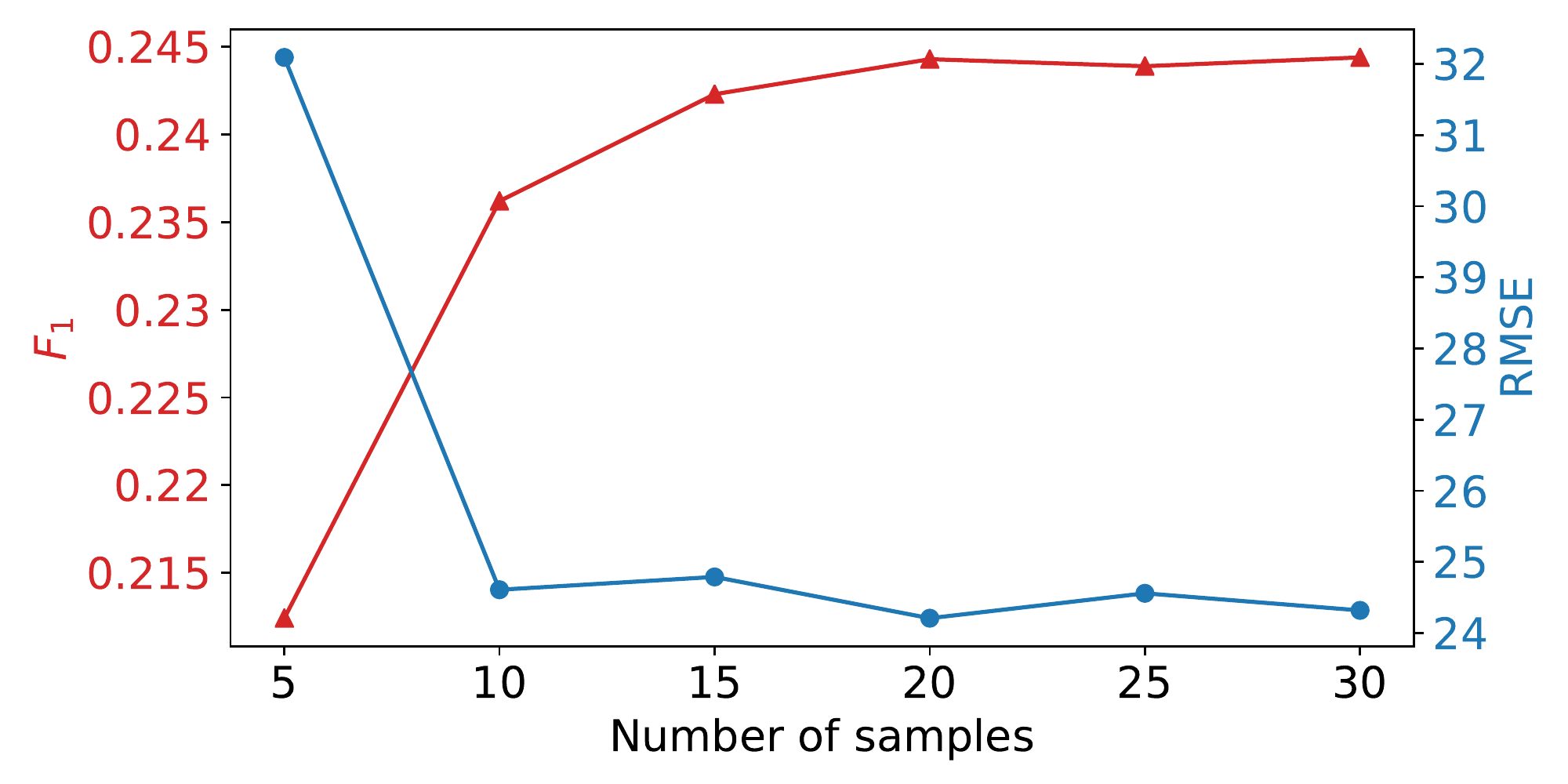}
	\vspace{-2em}
	\caption{The influence of the number of samples of the Monte Carlo estimation on SAHP's performance for event prediction.}
	\centering
	\label{fig:num_sample}
	\vspace{-2em}
\end{figure}

\paragraph{Number of samples' influence.}
When we optimize the objective function Eq.~\ref{eq:obj}, since it is not a closed form of the expectation, we use Monte Carlo sampling to approximate the integral. 
This experiment studies how the number of samples influences the SAHP's performance. 
The number of samples varies from 5 to 30 with step size 5. We report experimental results obtained from the StackOverflow dataset; other datasets share similar findings. 

Figure~\ref{fig:num_sample} describes how the performance of event prediction changes with different number of samples. From 5 to 10 samples, there is a significant improvement on the evaluation metrics. With more than 10 samples, we observe that the performance plateaus. 
To reduce computational time, we use 10 as the default number of samples in Monte Carlo.

\paragraph{Model interpretability.}

\begin{figure}[!ht]
	\centering
	\label{fig:atten}
    \includegraphics[width=0.48\textwidth]{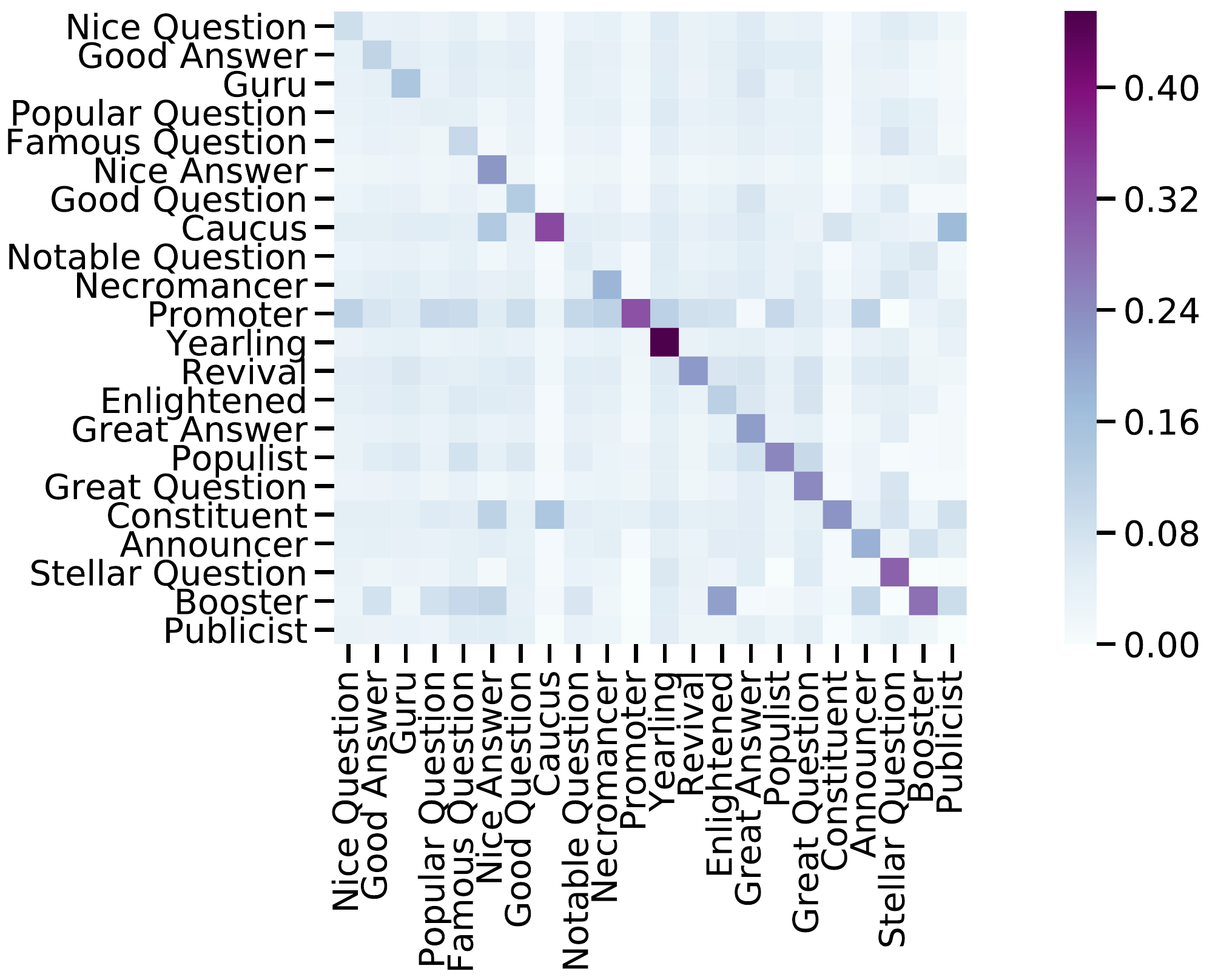}
    \vspace{-2em}
    \caption{Expected attention weights among event types on the StackOverflow test set.}
    \vspace{-2em}
\end{figure}

Apart from strong capacity in reconstructing the intensity function, the other advantage of our method is its higher interpretability. SAH is able to reveal peer influence among event types.
To demonstrate that, we extract the attention weight that the type-$u$ events allocate to the type-$v$ events and accumulate such attention weight over all the sequences on the StackOverflow test set. We remove the effect of the frequency of the ($u,v$) pairs in the dataset through dividing the accumulated attention weight via the ($u,v$) frequency. After normalization, we obtain the statistical attention distribution as shown in Figure~\ref{fig:atten}. The cell at the $u$-th row and $v$-th column means the statistical attention that the type-$u$ allocates to the type-$v$. 

Two interesting findings can be drawn from this figure: 1) for most cells in the diagonal line, when the model computes the intensity of one type, it attends to the history events of the same type; 2) for dark cells in the non-diagonal line, such as (\emph{Constituent, Caucus}), (\emph{Boosters and Enlightened}) and (\emph{Caucus and Publicist}), the model attends to the latter when computing the likelihood of the former. The first finding is attributed to the fact that attention is computed based on similarity between two embeddings while the second finding indicates the statistical co-occurrence of event types in a sequence.



\vspace{-0.5em}
\section{Related Work}
\label{sec:related}
\vspace{-0.5em}
\paragraph{Neural temporal point process.}

Complicated dynamics of event occurrence demands for higher capacity of Hawkes processes. To meet this demand, neural networks have been incorporated to modify the intensity function. 
~\citet{du2016recurrent} proposed a discrete-time RNN to encode history to fit parameters of the intensity function.
\citet{mei2017neural} designed a continuous-time Long Short-Term Memory model that avoids encoding time intervals as explicit inputs. Another two works chose not to model the intensity function. ~\citet{omi2019fully} proposed to model the cumulative distribution function with a feed forward neural network, yet it suffers from two problems: (1) the probability density function is not normalized and (2) negative inter-event times are assigned non-zero probability, as claimed by ~\citet{shchur2019intensity}. Then, ~\citet{shchur2019intensity} suggested modeling the conditional probability density distribution by a log-normal mixture model. They only studied the one-dimensional distribution of inter-event times, neglecting mutual influence among different event types. Also, despite claimed flexibility, it fails to achieve convincing performance on predicting event types. 
Above all, history has always been encoded by a recurrent structure.

However, RNN and its variants have been empirically proved to be less competent than self-attention in NLP~\citep{DBLP:conf/nips/VaswaniSPUJGKP17,devlin2018bert}.
Moreover, RNN-modified Hawkes processes do not provide a simple way to interpret the peer influence among events. Each historical event updates hidden states in the RNN cells but the process is lack of straightforward interpretability~
\citep{karpathy2015visualizing,krakovna2016increasing}. 

Besides,
recent works have advanced methods of parameter optimization. 
Alternatives of maximum likelihood estimation can be adversarial training~\citep{xiao2017wasserstein}, online learning~\cite{yang2017online}, Wasserstein loss~\citep{xiao2018learning}, noise contrastive estimation~\citep{guo2018initiator} and reinforcement learning~\citep{li2018learning,upadhyay2018deep}. This line of research is orthogonal to our work.


\vspace{-1em}
\paragraph{Position embedding.}
Self-attention has to rely on position embeddings to capture sequential orders. 
~\citet{DBLP:conf/nips/VaswaniSPUJGKP17} computed the absolute position embedding by feeding order numbers to sinusoidal functions. In contrast, the relative position embeddings use relative distance of the center token to others in the sequence. ~\citet{shaw2018self} represented the relative position by learning an embedding matrix. ~\citet{wang2019self} introduced a structural position to model a grammatical structure of a sentence, which involves both the absolute and the relative strategy. 
However, these methods only consider order numbers of tokens, which ignores time intervals for temporal event sequences.

\vspace{-0.5em}
\section{Conclusion}
\label{sec:conclusion}
\vspace{-0.5em}
The intensity function plays an important role in Hawkes processes for predicting asynchronous events in the continuous time domain. 
In this paper, we propose a self-attentive Hawkes process where self-attention is adapted to enhance the expressivity of the intensity function. This method enhances the model prediction and model interpretability. For the former, the proposed method outperforms state-of-the-art methods via better capturing event dependencies; while for the latter, the model is able to reveal peer influence via attention weights. For future work, we plan to extend this work for causality analysis of asynchronous events.

\newpage

\bibliographystyle{icml2020}

\clearpage
\appendix

\section{Optimization}
Given the history $\mathcal{H}(t_{i+1})=\{(v_1,t_1) ,\dots,(v_i,t_i) \}$, the time density of the subsequent event is calculated as: 
\begin{equation}
p_{i+1}(t)= P(t_{i+1}=t|\mathcal{H}(t_{i+1}))=\lambda(t)\exp\left({-\int_{t_i}^t \lambda(s)ds}\right),
\end{equation}
where $\lambda(t)=\sum_u \lambda_u(t)$. 
The prediction of the next event timestamp $t_{i+1}$ is equal to the following expectation:
\begin{equation}
\hat{t}_{i+1}=\mathbb{E}[t_{i+1}| \mathcal{H}(t_{i+1})]=\int_{t_i}^{\infty} t p_{t_{i+1}}(t)dt.
\end{equation}
While the prediction of the event type is equal to: \begin{equation}
    \hat{u}_{i+1}=\argmax_{u \in \mathcal{U}} \int_{t_i}^{\infty} \frac{\lambda_u(t)}{\lambda(t)}p_{i+1}(t)dt.
\end{equation}
Because this integral is not solvable analytically we approximate it via Monte Carlo sampling.

To learn the parameters of the proposed method, we perform a Maximum Likelihood Estimation (MLE). 
Other advanced and more complex 
adversarial learning~\citep{xiao2017wasserstein} and
reinforcement learning~\citep{li2018learning} methods have been proposed, however we use MLE for its simplicity. 
We use the same optimization method for our model and all baselines as done in their original papers. 
To apply MLE, we derive a loss function based on the negative log-likelihood.
The likelihood of a multivariate Hawkes process over a time interval $[0,T]$ is given by:
\begin{equation}
    \mathcal{L} (\lambda) = \sum^{L}_{i=1} \log \lambda_{v_{i}} (t_i) - \int_0^T \lambda(\tau)d\tau,
\end{equation}
where the first term is the sum of the log-intensity functions of past events, and the second term corresponds to the log-likelihood of infinitely many non-events. 
Intuitively, the probability that there is no event of any type in the infinitesimally time interval $[t,t+dt)$ is equal to $1-\lambda(t) dt$, the log of which is $-\lambda(t)dt$.


\section{Datasets}

\begin{figure*}
	\centering
	\includegraphics[width=0.85\textwidth]{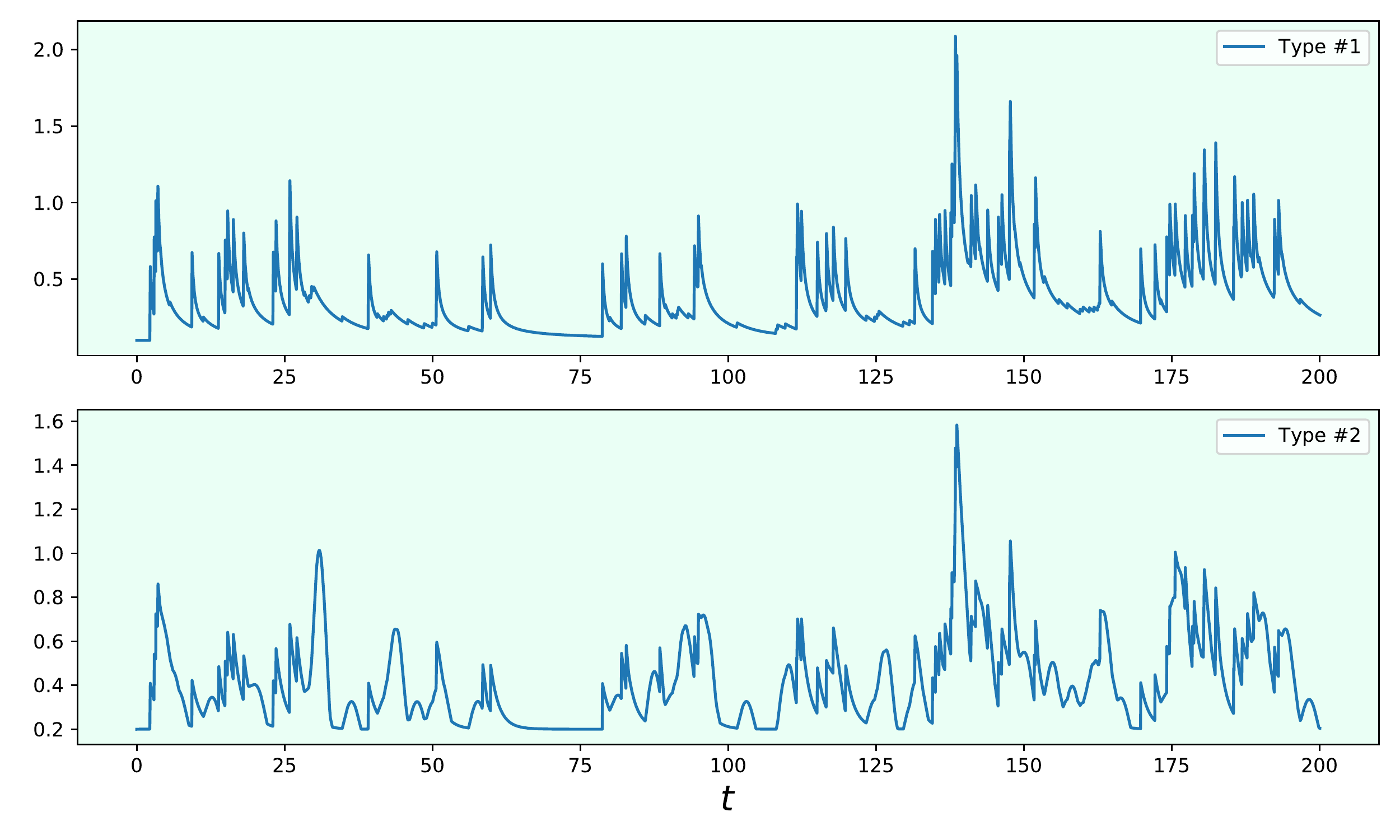}
	\caption{The intensities of the two-dimensional Hawkes processes over the synthetic dataset. The upper and the lower subfigure correspond to the dimension-$1$ and the dimension-$2$ respectively.	}
	\centering
	\label{fig:intensity_curve}
\end{figure*}


\paragraph{Retweet (RT)} 
The Retweet dataset contains a total number of 24,000 retweet sequences. In each sequence, an event is a tuple of the tweet type and time. 
There are $U=3$ types: ``small'', ``medium'' and ``large'' retweeters. 
The ``small'' retweeters are those who have fewer than 120 followers, ``medium'' retweeters have more than 120 but fewer than 1,363 followers, and the rest are ``large'' retweeters. 
As for retweet time, the first event in each sequence is labeled with 0, the next events are labeled with reference to their time interval with respect to the first event in this sequence. The dataset contains information of when a post will be retweeted by which type of users.

\paragraph{StackOverflow (SOF)}

The StackOverflow dataset includes sequences of user awards within two years. StackOverflow is a question-answering website where users are awarded based on their proposed questions and their answers to questions proposed by others. 
This dataset contains in total 6,633 sequences. There are $U=22$ types of events: Nice Question, Good Answer, Guru, Popular Question, Famous Question, Nice Answer, Good Question, Caucus, Notable Question, Necromancer, Promoter, Yearling, Revival, Enlightened, Great Answer, Populist, Great Question, Constituent, Announcer, Stellar Question, Booster and Publicist. The award time records when a user receives an award. With this dataset, we can learn which type of awards will be given to a user and when.

\paragraph{MIMIC-II (MMC)}

The Multiparameter Intelligent Monitoring in Intensive Care (MIMIC-II) dataset is developed based on an electric medical record system. The dataset contains in total 650 sequences, each of which corresponds to an anonymous patient's clinical visits in a seven-year period. Each clinical event records the diagnosis result and the timestamp of that visit.
The number of unique diagnosis results is $U=75$. According to the clinical history, a temporal point process is supposed to capture the dynamics of when a patient will visit doctors and what the diagnose result will be.



\end{document}